\pdfoutput=1
\documentclass{article} 
\usepackage{nips15submit_e,times}
\usepackage{hyperref}
\usepackage{url}
\usepackage{graphicx}
\usepackage{caption}
\usepackage{subcaption}
\usepackage{amsmath}
\usepackage{caption}
\usepackage{subcaption}
\usepackage{todonotes}
\usepackage{hhline}

\title{Improved Deep Learning Baselines for Ubuntu Corpus Dialogs}

\author{
Rudolf Kadlec, Martin Schmid and Jan Kleindienst \\
IBM Watson\\
V Parku 4, Prague 4, Czech Republic\\
\texttt{\{rudolf\_kadlec, martin.schmid, jankle\}@cz.ibm.com} \\
}

\newcommand{\RUDA}[1]{{\color{black}#1}} 

\newcommand{\MARTINSECOND}[1]{{\color{black}#1}} 
\newcommand{\MARTINTHIRD}[1]{{\color{black}#1}} 

\nipsfinalcopy 

\begin{document}

\maketitle

\begin{abstract}
\MARTINSECOND{
This paper presents results of our experiments for the next utterance ranking on the Ubuntu Dialog Corpus -- the largest publicly available multi-turn dialog corpus.
First, we use an in-house implementation of previously reported models to do an independent evaluation using the same data.
Second, we evaluate the performances of various LSTMs, Bi-LSTMs and CNNs on the dataset.
Third, we create an ensemble by averaging predictions of multiple models. The ensemble further improves the performance and it achieves a state-of-the-art result for the next utterance ranking on this dataset.
Finally, we discuss our future plans using this corpus.
}
\end{abstract}

\section{Introduction}


\MARTINSECOND
{
The Ubuntu Dialogue Corpus is the largest freely available multi-turn based dialog corpus ~\cite{lowe2015ubuntu}\footnote{\url{http://cs.mcgill.ca/~jpineau/datasets/ubuntu-corpus-1.0/}}.
It was constructed from the Ubuntu chat logs\footnote{\url{http://irclogs.ubuntu.com/}} --- a collection of logs from Ubuntu-related chat rooms on the Freenode IRC network.
Although multiple users can talk at the same time in the chat room, the logs were preprocessed using heuristics to create two-person conversations.
The resulting corpus consists of almost one million two-person conversations, where a user seeks help with his/her Ubuntu-related problems
(the average length of a dialog is $8$ turns, with a minimum of 3 turns).
Because of its size, the corpus is well-suited for explorations of deep learning techniques in the context of dialogue systems.
In this paper, we introduce our preliminary research and experiments with this corpus, and report state-of-the-art results.

The rest of the paper continues as follows: 
1. we introduce the setup --- the data as well as the evaluation of the task;
2. we briefly describe the previously evaluated  models;
3. we introduce three different models (one of them being the same as in the previous work);
4. we evaluate these models and experiment with different amount of training data;
5. we conclude and discuss our plans for future works
}




\MARTINSECOND
{
\section{Data}
In this section we briefly describe the data and evaluation metrics used in~\cite{lowe2015ubuntu}. 
First, all the collected data was preprocessed by replacing named entities with corresponding
tags (name, location, organization, url, path). This is analogical to the prepossessing of \cite{vinyals2015neural} 
(note that the IT helpdesk dataset used there is not publicly available).
Second, these data are further processed to create tuples of $(context, response, flag)$.
The $flag$ is a Boolean variable indicating whether the response is correct
 or incorrect.

To form the training set, each utterance (starting from the third one) is considered as a
potential response, while the previous utterances form its context.
So a dialogue of length $n$ yields $(n-2)$ training examples $(context, response, 1)$ and 
$(n-2)$ training examples $(context, response', 0)$.
The negative response $response'$ is a randomly sampled utterance from the entire corpus. 
Finally, the training examples are shuffled.




\subsection{Evaluation Metric}
A randomly selected $2\%$ of the conversations are used to create a test set.
The proposed task is that of the best response selection.
The system is presented with $n$ response candidates, and it is asked to rank them.
To vary the task's difficulty (and to remedy that some of the sampled candidates flagged as incorrect can very well be correct), 
the system's ranking is considered correct if the correct response is among the first $k$ candidates. This quantity is denoted as Recall@k.
The baselines were reported with $(n,k)$ of $(2,1)$, $(10,1)$, $(10,2)$ and $(10,5)$.
}

\section{Approaches}

This task can naturally be formulated as a ranking problem which is often tackled by three techniques~\cite{liu2009learning}: (i) pointwise; (ii) pairwise and (iii) listwise ranking.

While pairwise and listwise ranking approaches are empirically superior to the pointwise ranking approach, our preliminary experiments use pointwise ranking approach for its simplicity. Note that pointwise method was also used in the original baselines~\cite{lowe2015ubuntu}.

\subsection{Pointwise Ranking}

In pointwise ranking, only the context and the response are directly used to compute the probability of the pair. All the pairs are then sorted by their probabilities. 
We denote the function that outputs the probability of the pair as $g(context, response)$.
In our settings, the function $g$ is represented by a neural network (learned using the training data).
We describe the details of the network architectures used in the following sections.

\MARTINSECOND
{
\section{Previous Work}


The pointwise architectures reported in \cite{lowe2015ubuntu} included (i) TF-IDF, (ii) RNN and (iii) LSTM.
In this section, we briefly describe these models. 

\begin{figure}[h!]
  \centering
  \includegraphics[width=4in]{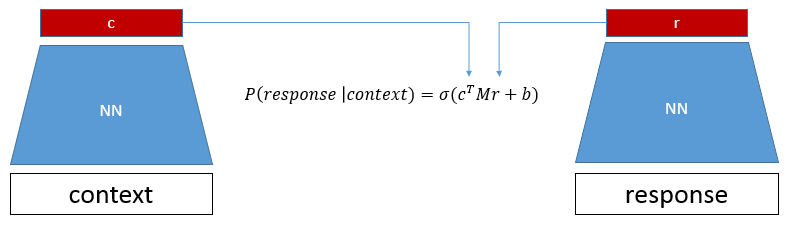}
  \caption   
  {
  Neural Network Embedding approach. 
  A neural network is used to compute the embedding for the context and the response, denoted as $c$ and $r$.
  These are fed through a sigmoid function to compute the pairwise probability.
  }
  \label{fig-prevwork}
\end{figure}

\subsection{TF-IDF}
The motivation here is that the correct response tends to share more words with the context than the incorrect ones.
First, the TF-IDF vectors are calculated for the context and each of the candidate responses.
Next, the cosine similarity between the context vector and each response vector is used to rank the responses.

\begin{align}
    tfidf_{context}(w) &= tf(w,context) \times idf(w,D) \\
    tfidf_{document}(w) &= tf(w,document) \times idf(w,D) \\
    g(context, response) &= tfidf_{context} \cdot tfidf_{context}
\end{align}


\begingroup\vspace*{-\baselineskip}
{
    \emph{
    $tfidf_{context}$ and $tfidf_{response}$ are the resulting TF-IDF vectors for context and response respectively.
    $D$ stands for the corpus and $w$ is a word.
    The dimension of the resulting vectors is thus equal to the dictionary size.
    }
}
\vspace*{\baselineskip}\endgroup

\subsection{Neural Network Embeddings}
A neural network is used to create an embedding of both the context and the candidate response.
These embeddings, denoted as $c$ and $r$, are then multiplied using a matrix $M$ and the result is fed into the sigmoid function
to score the response.


\begin{align}
c &= f(context)\\
r &= f(response) \\
g(context, response) &= \sigma(c^{\top} M r + b)
\end{align}

\begingroup\vspace*{-\baselineskip}
{
    \emph{
    $c$ and $r$ are the resulting embeddings of the context and response, computed using a neural network.
    We present some different architectures to compute these embeddings.
    }
}
\vspace*{\baselineskip}\endgroup

Figure~\ref{fig-prevwork} illustrates the approach. Note that matrix $M$, bias $b$ and parameters of the function $f$ (which is a neural network) are all learned using the training data.

One can think of this approach as a predictive one --- given the context,
we predict the embedding of the response as
$r' = c^{\top}M$, and measure the similarity of the predicted response $r'$ to the actual response $r$ using the dot product
(or vice-versa, predicting the context from the response as $c' = Mr$)

The authors experimented with vanilla RNN and LSTM~\cite{Hochreiter1997} as the underlying networks producing the embeddings.
LSTM significantly outperformed RNN in the author's experiments.
}

\section{Our Architectures}

\MARTINSECOND
{

All our architectures fall within the neural network embedding based approach.
We implemented three different architectures
(i) CNN~\cite{kim2014convolutional} (ii) LSTM and (iii) Bi-Directional~\cite{Schuster1997} LSTM.
We also report an ensemble of our models.

All of our architectures share the same design where the words from the input sequence (context or response) are projected
into the words' embeddings vectors.
Thus, if the input sequence consist of $42$ words, we project these words into a matrix $E$ which has
a dimension $e \times 42$, where $e$ is dimensionality of the word embeddings.
}
\subsection{CNN}

\MARTINSECOND
{ 
While originating from computer vision \cite{lecun1998gradient}, CNN models have
recently been very successfully applied in NLP problems \cite{kim2014convolutional}.
At the very heart of the CNN model, the convolving filters are sequentially applied over the input sequence.
The width of the filters might vary, and in NLP typically range from $1$ to $5$
(the filters can be thought of here as a form of n-grams).
These filters are followed by a max-pooling layer to get a fixed-length input.
In our architecture, the output of the max-pooling operation forms the context/response embedding.
Thus, the resulting embedding has a dimension equal to the number of filters.
Figure \ref{fig:cnn} displays this architecture with two filters.
}


\subsection{LSTM}
\MARTINTHIRD{
Long short-term memory (LSTM) is a recurrent neural network (RNN) architecture designed to remedy the vanishing gradient problem of vanilla RNN~\cite{Hochreiter1997}.
Thus, LSTM networks are well-suited for working with (very) long sequences~\cite{Gers2002}.
}
We use the same model as the authors' LSTM network~\cite{Lowe2015}.
LSTM iterates over the sequence embeddings, and the resulting embedding is the last state of the LSTM's cells.
Figure \ref{fig:lstm} illustrates this architecture.


\subsection{Bi-Directional LSTM}
\begin{figure}
    \centering
    \begin{subfigure}[b]{0.35\textwidth}
        \includegraphics[width=\textwidth]{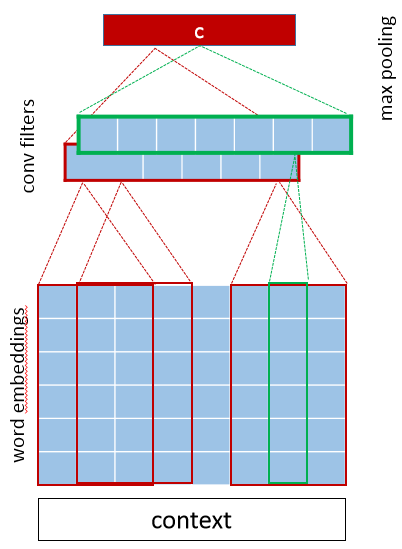}
        \caption{CNN}
        \label{fig:cnn}
    \end{subfigure}
    ~ \quad 
    \begin{subfigure}[b]{0.2\textwidth}
        \includegraphics[width=\textwidth]{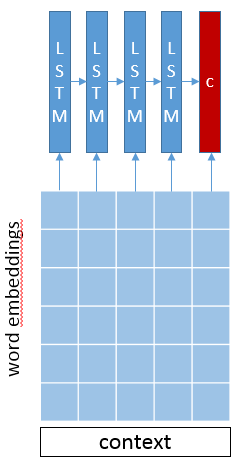}
        \caption{LSTM}
        \label{fig:lstm}
    \end{subfigure}
    ~ \quad 
    \begin{subfigure}[b]{0.2\textwidth}
        \includegraphics[width=\textwidth]{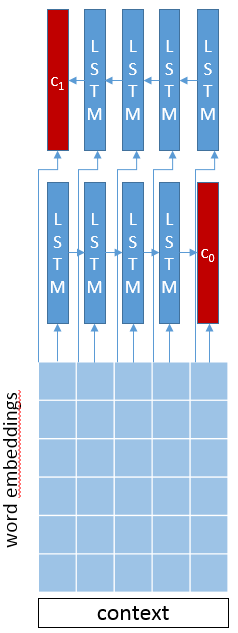}
        \caption{Bi-Directional}
        \label{fig:bidirect}
    \end{subfigure}
    \caption{
    Different architectures to compute the embedding of the context/reponse:
    a) CNN with two feature maps: the green one has feature width one, the red one has feature width three.
        Since the output out of the maxpooling forms our context embedding, the resulting embedding has dimension equal to the number of filters ($2$ in this example)
    b) LSTM network. Embeddings is the last hidden state, thus the dimension of the embedding equals to the number of LSTM units.
    c) Bi-Directional network. Embedding is a concatenation of the corresponding LSTM states: $c = c1.c2$
    }
    \label{fig:animals}
\end{figure}

Although the LSTM is tailor-made to keep context over large sequences,
empirically it can be problematic for the network to capture the meaning of the entire sequence as it gets
longer.
If the important parts of the sequence are found at the beginning of a long sequence, the LSTM might struggle to get well-performing embedding.
We decided to experiment with Bi-LSTMs to see whether this is the case in our settings.
Bi-directional~\cite{Schuster1997} LSTMSs feed the sequence into two recurrent networks --- one reads the sequence as it is, the second reads the sequence from the end to the beginning.
To avoid forming cycles, only the outputs of the recurrent networks (not the state-to-state connections) lead to same units in the next layers.
Figure \ref{fig:bidirect} illustrates this architecture.

\section{Experiments}

\subsection{Method}
\RUDA{
To match the original setup of~\cite{lowe2015ubuntu} we use the same training data\footnote{The dataset in binary format is available at \url{http://cs.mcgill.ca/~jpineau/datasets/ubuntu-corpus-1.0/ubuntu_blobs.tgz} [accessed 25.9.2015]}. We use one million training examples and we use the same word vectors pre-trained by GloVe~\cite{Pennington2014a}. All our models were implemented using Theano~\cite{Bastien-Theano-2012} and Blocks~\cite{VanMerrienboer2015}. 
For training we use ADAM learning rule~\cite{Kingma2015} and binary negative log-likelihood as training objective. We stop the training once Recall@1 starts increasing on a validation set. The experiments were executed on Nvidia K40 GPUs. The best meta-parameters were found by simple grid search. 

In all architectures we tried both: (i) learning separate parameters for the networks encoding context and response and (ii) learning shared parameters for both networks. Here we report only the results for the architectures with shared parameters, since they consistently achieved higher accuracy.

Aside from learning single models, we also experimented with model ensembles. We found that averaging predictions of multiple models further improves performance, which is common in many machine learning tasks~\cite{IlyaSutskeverOriolVinyals2014,mesnil2015}. Our best classifier is an ensemble of 11 LSTMs, 7 Bi-LSTMs and 10 CNNs trained with different meta-parameters. 
}
\subsection{Results}

Table \ref{tbl-results} shows performance of the models with the best metaparameters in each category.  An example prediction from the ensemble is shown in Table~\ref{tbl-dialog}.
\RUDA{The performance was evaluated after every epoch of training. Most of the models achieved the best cost on validation data after a single epoch of training. However, the best Recall metrics were usually recorded in the second epoch of training. 
}

\begin{table}[h!]
\centering
\begin{tabular}{l||l|l|l||l|l|l||l|}
\cline{2-8}
                                  & \multicolumn{3}{l||}{Baselines from~\cite{lowe2015ubuntu}} & \multicolumn{4}{l|}{Our Architectures} \\ \cline{2-8} 
                                  & TF-IDF      & RNN         & LSTM        & CNN       & LSTM       & Bi-LSTM & Ensemble      \\ \hline
\multicolumn{1}{|l||}{1 in 2 R@1}  & 65.9\%      & 76.8\%      & 87.8\%      &   84.8\%           & 90.1\%           & 89.5\%  & \textbf{91.5\%}               \\ \hline
\multicolumn{1}{|l||}{1 in 10 R@1} & 41.0\%      & 40.3\%      & 60.4\%      & 54.9\%          & 63.8\%           & 63.0\%   & \textbf{68.3\%}          \\ \hline
\multicolumn{1}{|l||}{1 in 10 R@2} & 54.5\%      & 54.7\%      & 74.5\%      & 68.4\%          & 78.4\%           & 78.0\%   & \textbf{81.8\%}        \\ \hline
\multicolumn{1}{|l||}{1 in 10 R@5} & 70.8\%      & 81.9\%      & 92.6\%      & 89.6\%          & 94.9\%           & 94.4\%    & \textbf{95.7\%}         \\ \hline
\end{tabular}
\caption{\RUDA{Results of our experiments compared to the results reported in~\cite{lowe2015ubuntu}. Meta-parameters of our architectures are the following: our CNN had 400 filters of length 1, 100 filters of length 2 and 100 filters of length 3; our LSTM had 200 hidden units and our bidirectional LSTM had 250 hidden units in each network. For CNNs and LSTMs, the best results were achieved with batch size 256. For Bi-LSTM, the best batch size was 128.}}
\label{tbl-results}
\end{table}

\begin{table}[h!]
\centering
\begin{tabular}{c|l|p{11cm}}

Turn & User & Text \\ \hline
1	& A: &	 anyone know why " aptitude update " returns a non-successful status (255) ? \\		
2	& B: &	  does apt-get update work ? \\		
3	& A: &	  i ' ve been missing updates because my normal process is sudo bash -c " aptitude update \&\& aptitude safe-upgrade -y ". ahh , " e : some index files failed to download . they have been ignored , or old ones used instead .". so i guess the issue is that " aptitude update " is n't giving an error at all \\		
\end{tabular}

\vskip1\baselineskip
\begin{tabular}{c|l|c|p{8cm}}
N-Best	& \multicolumn{2}{c|}{Confidence} & Response\\		\hline	
1	& 	 \textbf{******} &	\textbf{0.598}	& \textbf{does the internet work on that box ?} \\
2	&  	 **** &	0.444	& what time is it saying to going to be released ?? \\
3	&  	 *** &	0.348	& ahh ok \\
4	&  	 ** &	0.245	& nice \\
\end{tabular}
\caption{\RUDA{A dialog context with three turns and a set of four ranked possible responses. The highest ranked response is the ground truth response in this case.}}
\label{tbl-dialog}
\end{table}

\subsection{Discussion}

Our ensemble of classifiers sets a new state-of-the art performance for response ranking on the Ubuntu Dialog Corpus --- the largest, publicly available multi-turn dialog corpus. Interestingly LSTMs and Bi-LSTMs achieve almost the same accuracy. We hypothesise that: (i) either utterances that appear at the beginning of the context are less important than the later utterances or, (ii) LSTMs successfully capture all of the important parts of the sequence. When we inspect accuracy of individual models we see that recurrent models are superior to CNNs. However, CNNs proved to significantly improve performance of the ensemble. An ensemble without the 10 CNNs had Recall@1 accuracy of only 66.8 compared to 68.3 of the larger ensemble. This shows that CNNs learned representations that are complementary to the recurrent models.
We believe that our results are important, since they can be used as baselines for more complicated models (see the Future Work).

\subsection{Varying Training Data Size}
\begin{figure}[h!]
  \centering
  \includegraphics[width=5in]{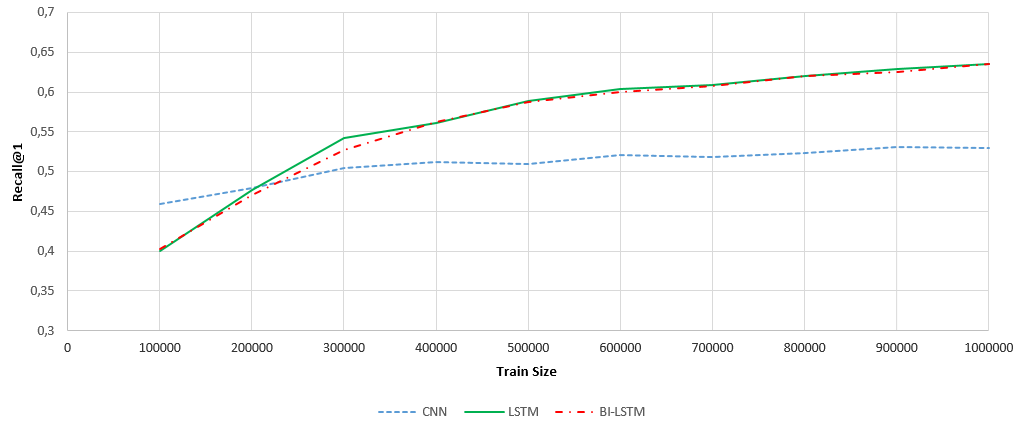}
  \caption   
  {
  Training data size ranging from $100,000$ to the full $1,000,000$ examples ($X$ axis) and the resulting
  Recall@1 ($Y$ axis). The CNN has 500, 100 and 100 filters of length 1, 2 and 3. The LSTM and Bi-LSTM has both 300 hidden units in each recurrent layer.
  }
  \label{fig-graph}
\end{figure}

\MARTINSECOND{
We also experimented with different training data sizes in order to see how this affects the resulting models.
We trained all networks on a training data size ranging from $100,000$ to the full $1,000,000$ examples.
The graph in Figure \ref{fig-graph} shows the Recall@1 for all the three models (reported on the test data).
There are two main observations here: 
(i) CNNs outperform recurrent models if the training dataset is small.
We believe that this is mostly due to the \textit{max} operation performed on top of the feature maps.
Thanks to the simplicity of this operation, the model does not over-fit the data and generalizes better when learned on small training datasets.
On the other hand, the simplicity of the operation does not allow the model to properly handle more complicated dependencies 
(such as the order in which the n-grams occur in the text), thus recurrent models perform better given enough data;
(ii) the recurrent models have not made its peak yet, suggesting that adding more training data would improve the model's accuracy.
This agrees with Figure 3 of the previous evaluation \cite{lowe2015ubuntu}.
}

\section{Future Work}
In our future work, we plan to investigate applicability of neural networks architectures extended with memory (e.g.,~\cite{Graves2014,Sukhbaatar2015,Joulin2015}) on this task.
It is an appealing idea to bootstrap the system with external source of information (e.g., user manual or man pages)
to help the system pick the right answer. For successful application of this paradigm in the domain of reinforcement learning, see~\cite{Branavan2012}.

An alternative direction for future research might be to extend the model with attention~\cite{Bahdanau2014} over sentences in the dialog context. This would allow the model to explain which facts in the context were the most important for its prediction. Therefore, the prediction could be better interpreted by a human. 

Additional accuracy improvements might be also achieved by different text pre-processing pipelines. For instance, in the current dataset all named entities were replaced with generic tags, which could possibly harm the performance. 

\section{Conclusion}
\RUDA{
In this work we achieved a new state-of-the-art results on the next utterance ranking problem recently introduced in~\cite{lowe2015ubuntu}. The best performing system is an ensemble of multiple diverse neural networks. In the future, we plan to use our system as a base for more complicated models going beyond the standard neural network paradigm.
}






\bibliographystyle{ieeetr}
\bibliography{paper}

\end{document}